\documentclass[conference]{IEEEtran}
\IEEEoverridecommandlockouts
\usepackage{cite}
\usepackage{amsmath,amssymb,amsfonts}
\usepackage{algorithmic}
\usepackage{graphicx}
\usepackage{textcomp}
\usepackage{xcolor}
\usepackage{booktabs}
\usepackage{makecell}
\usepackage{hyperref}
\def\BibTeX{{\rm B\kern-.05em{\sc i\kern-.025em b}\kern-.08em
    T\kern-.1667em\lower.7ex\hbox{E}\kern-.125emX}}

\begin{document}

\title{Fine-Grained Classification: Connecting Metadata via Cross-Contrastive Pre-Training}

\author{\IEEEauthorblockN{Sumit Mamtani}
\IEEEauthorblockA{\textit{New York University} \\
sm9669@nyu.edu}
~\\
\and
\IEEEauthorblockN{ Yash Thesia}
\IEEEauthorblockA{\textit{New York University} \\
yt2188@nyu.edu}
}

\maketitle

\begin{abstract}
 Fine-grained visual classification aims to recognize objects belonging to many subordinate categories of a supercategory, where appearance alone often fails to distinguish highly similar classes.
We propose a unified framework that integrates \emph{image}, \emph{text}, and \emph{metadata} via \emph{cross-contrastive pre-training}.
We first align the three modality encoders in a shared embedding space and then fine-tune the image and metadata encoders for classification.
On \textbf{NABirds}~\cite{NAbird}, our approach improves over the baseline by \textbf{7.83\%} and achieves \textbf{84.44\%} top-1 accuracy, outperforming strong multimodal methods.
\end{abstract}
\begin{IEEEkeywords}
Fine-Grained Classification, Contrastive Learning, CNN, DistilBERT, Geo-prior
\end{IEEEkeywords}

\section{Introduction}

Fine-grained visual classification (FGVC) separates instances within a basic category into visually similar sub-categories (e.g., bird species). The challenge is that inter-class differences are subtle while intra-class variation (pose, background, lighting) can be large. Vision–language encoders trained at scale have helped transfer to fine-grained tasks as image–text pre-training continues to mature; for instance, SigLIP~2 combines captioning-based pre-training, self-distillation, masked prediction, and online data curation to produce stronger image–text features than earlier CLIP-style models~\cite{SigLIP2_2025}.

A complementary line of work shows that \emph{where} and \emph{when} an image was captured can be just as informative as appearance. Aligning location encoders to images (e.g., GeoCLIP) embeds geospatial structure directly into the representation~\cite{GeoCLIP_2023}, and recent methods learn multi-resolution geo-embeddings with strong transfer across tasks and datasets~\cite{RANGE_CVPR_2025}. In ecology and biodiversity monitoring, incorporating spatio-temporal context consistently improves species mapping and identification~\cite{Brun_NatComm_2024}, and concurrent analyses report that explicit geo priors can boost species-level FGVC~\cite{GeoPriors_2025}.

\textbf{This paper.} We propose a cross-contrastive pre-training framework that brings images, class text, and spatio-temporal metadata into a single 256-D embedding space. Concretely, we encode GPS and date with sinusoidal features followed by a small MLP, project all three modalities into the shared space, and optimize a six-term contrastive objective that aligns \emph{every} pair in both directions (image$\leftrightarrow$text, image$\leftrightarrow$metadata, text$\leftrightarrow$metadata). After pre-training, we discard the text branch and fine-tune a lightweight classifier on the concatenated image+metadata embedding. On NABirds~\cite{NAbird}, the approach reaches \textbf{84.44\%} top-1 accuracy (\textbf{+7.83\%} over our vision-only baseline), suggesting that coupling appearance with spatio-temporal context helps disambiguate look-alike species whose ranges differ by region or season.

\noindent\textbf{Contributions.}
\begin{itemize}
    \item \textbf{Tri-modal alignment for FGVC.} A unified framework that models image appearance, geospatial/date metadata, and class text in one embedding space for fine-grained recognition.
    \item \textbf{Cross-contrastive objective.} A six-term class-positive contrastive loss that aligns each modality pair in \emph{both} directions (image$\leftrightarrow$text, image$\leftrightarrow$metadata, text$\leftrightarrow$metadata), improving transfer over two-term image–text objectives~\cite{SigLIP2_2025}.
    \item \textbf{Simple fine-tuning head.} A small two-layer classifier on the concatenated image+metadata embeddings yields strong results with minimal overhead.
    \item \textbf{Results on NABirds.} \textbf{84.44\%} top-1 accuracy on NABirds~\cite{NAbird}, a \textbf{+7.83\%} gain over our vision-only baseline.
\end{itemize}

\section{Related Work}

\noindent\textbf{Vision-only fine-grained classification.}
Early FGVC work improved recognition along two complementary fronts: (i) \emph{localizing} subtle, part-level cues~\cite{multiattention-CNN,Ding_2019_ICCV,zheng2019looking,LearningtoNavigate,attentive-pairwise-interaction} and (ii) \emph{learning} feature representations that accentuate fine-grained differences while down-weighting nuisance factors such as pose, background, and illumination~\cite{BilinearCNNs,bilinear-p0oling,deep-bilinear-transformation}. Older pipelines leaned on explicit part detectors and attribute supervision~\cite{weakly-complementary-models}; more recently, strong backbones and supervised contrastive objectives trained at scale have become standard practice~\cite{Reset50,ContrastiveSupervised}. Even so, models that rely on appearance alone still struggle with sister species whose plumage, shape, or coloring are nearly indistinguishable—especially under difficult viewpoints or lighting—which motivates bringing in signals beyond pixels.

\noindent\textbf{Vision--language pre-training.}
Large image--text encoders align images with natural language and transfer well to fine-grained tasks~\cite{CLIP,combine-vision-language}. Ongoing work continues to refine these recipes; for example, SigLIP~2 combines captioning-style learning, self-distillation, masked prediction, and online data curation to strengthen image--text features over CLIP-style models~\cite{SigLIP2_2025}. However, text alone does not encode \emph{where} and \emph{when} a photo was taken, and short class prompts often miss ecological or seasonal context that helps separate look-alike taxa.

\noindent\textbf{Geospatial and metadata priors.}
A complementary thread incorporates location and time as priors. GeoCLIP aligns location encoders with image features so that geospatial structure is captured directly in the embedding space~\cite{GeoCLIP_2023}, while newer approaches learn multi-resolution geo-embeddings with strong transfer across datasets and tasks~\cite{RANGE_CVPR_2025}. In ecology and biodiversity monitoring, spatio-temporal context reliably improves species mapping and identification~\cite{Brun_NatComm_2024}, and concurrent analyses show that explicit geo priors can boost species-level FGVC~\cite{GeoPriors_2025}. Beyond raw GPS and date, prior knowledge such as ecoregions, elevation bands, or seasonality calendars can be folded into the metadata encoder to provide useful locality and periodicity biases.

\noindent\textbf{Multimodal fusion for FGVC.}
Many multimodal systems \emph{inject} metadata either early (as extra channels or conditioning inside the vision backbone) or late (via feature concatenation near the classifier), and some combine separate predictors with a learned prior~\cite{weakly-complementary-models}. These strategies can help, but they neither ensure that metadata is semantically aligned with visual/text cues nor guarantee that it shapes the representation geometry consistently across classes. Our approach instead \emph{aligns} image, text, and metadata in a shared space \emph{before} classification using a six-term cross-contrastive objective (image$\leftrightarrow$text, image$\leftrightarrow$metadata, text$\leftrightarrow$metadata), and then applies a lightweight head to the concatenated image+metadata embedding. This tri-modal alignment couples spatio-temporal context with visual and textual cues, helping to separate look-alike species that chiefly differ by range or season.

\noindent\textbf{Positioning and compatibility.}
Relative to two-term image--text objectives, our design ties metadata to \emph{both} image and text, encouraging the model to resolve visually ambiguous classes using geo-temporal context. The objective is model-agnostic and can pair with recent vision--language encoders and geo-embedding methods (e.g., SigLIP~2~\cite{SigLIP2_2025}, GeoCLIP~\cite{GeoCLIP_2023}, RANGE~\cite{RANGE_CVPR_2025}) without changing the loss.

\section{Proposed Model}
\begin{figure*}
  \centering
\includegraphics[width=\linewidth]{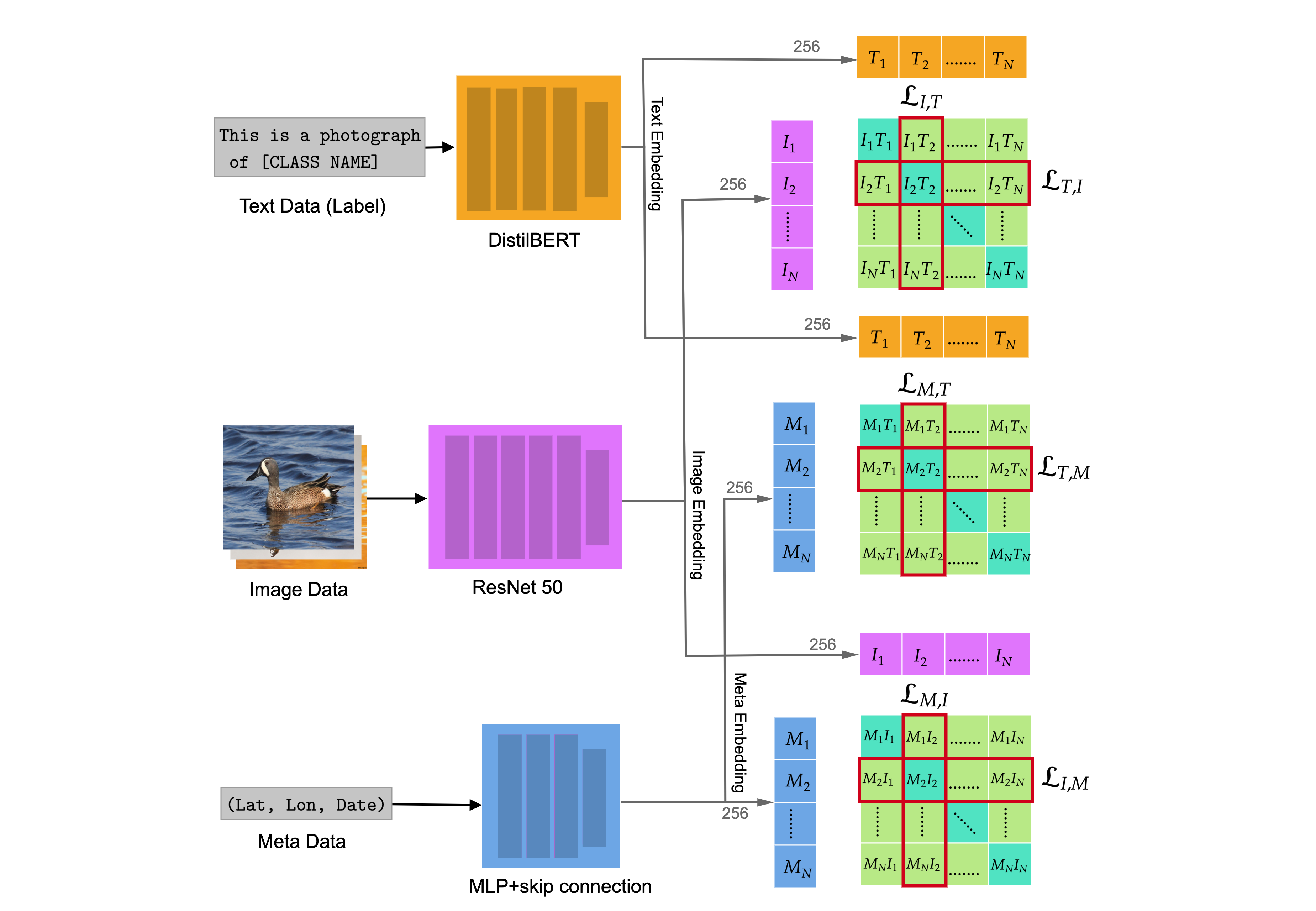}
  \caption{Proposed architecture—pre-training of text, metadata, and image encoders using cross-contrastive loss.}
  \label{fig:pre-training}
\end{figure*}
In this section, we discuss the proposed model architecture for fine-grained classification.
We train the model in two steps: First, we train the model using cross-contrastive pre-training, which is inspired by the CLIP\cite{CLIP} model. Then, we use the embeddings of image and metadata encoder as input to a shallow fully connected layer for the classification of 555 classes\cite{NAbird}.
\subsection{Cross-Contrastive Pre-Training}
Fig.~\ref{fig:pre-training} illustrates our pre-training stage. The objective aligns embeddings across modalities for samples that share the same class label.
\textbf{Image Embedding}: We use a ResNet-50 pre-trained on ImageNet~\cite{Reset50} and append a linear head to project 2048-D features to a 256-D embedding.
\textbf{Meta Embedding}: We first convert longitude, latitude, and date into multi-frequency sine--cosine features and feed them to a small MLP with a residual skip. The metadata encoder outputs a 256-D embedding. We then compute three $B\times B$ cosine-similarity matrices (image--text, image--metadata, text--metadata) and optimize both directions for each pair, yielding six losses with temperature $\tau$.
\textbf{Text Embedding}: We use the prompt ``This is a photograph of a bird called [CLASS]''~\cite{CLIP}, encode it with DistilBERT~\cite{DistilBert}, and project the 1024-D output to 256-D.
\textbf{Loss Function}: For each batch B, we compute the embedding of image, text, and metadata from their corresponding encoder models. Then, we project those embeddings to a shared 256-D space and normalize them. We utilized these outputs to compute three correlation matrices of $B\times B$\cite{CLIP} to find the similarity between each of the vector pairs. Corresponding to the computed matrix, we also calculated the label matrix. We put label 1 when the label corresponding to both embeddings are the same, otherwise zero

\noindent\textit{Normalization:} we use $\ell_2$-normalized embeddings in the loss,
\begin{equation}
\label{eq:norm}
\mathbf{z} \leftarrow \tilde{\mathbf{z}} / \|\tilde{\mathbf{z}}\|_2 \quad \text{for image, text, and metadata encoders.}
\end{equation}
\begin{align}
\mathcal{L}_{\text{total}} \;=\;&
\mathcal{L}_{I,T} + \mathcal{L}_{T,I} + \mathcal{L}_{M,T} + \mathcal{L}_{T,M} + \mathcal{L}_{M,I} + \mathcal{L}_{I,M} \label{eq:total} \\[0.25em]
\mathcal{L}_{I,T} \;=\;&
-\frac{1}{|I|} \sum_{i\in I} \frac{1}{|P(i)|} \sum_{p\in P(i)} \log
\frac{\exp\!\left( \mathbf{z}_i \cdot \mathbf{z}_p / \tau \right)}
{\sum_{a\in T} \exp\!\left( \mathbf{z}_i \cdot \mathbf{z}_a / \tau \right)}. \label{eq:it}
\end{align}

\noindent The remaining five terms are defined analogously.

\noindent\textit{where} $P(i) = \{\, p \in T \;|\; y_T(z_p)=y_I(z_i) \,\}$, 
Here, $|I|$ denotes the number of image samples in the minibatch.
$I$ is the batch of image embeddings, $T$ is the batch of text embeddings, and $M$ is the batch of metadata embeddings.
\paragraph*{\textbf{Intuition}}
The six cross-modal directions (image$\leftrightarrow$text, image$\leftrightarrow$metadata, text$\leftrightarrow$metadata) form a closed triangle of constraints. If image and text agree on class semantics and image and metadata agree on geo-temporal context, then text and metadata are also pulled into agreement. This transitive consistency distributes learning signals across modalities and yields compact, context-aware clusters while preserving the original loss design.

For each modality pair, we compute losses in both directions (row-wise and column-wise). Then, we back-propagate the total loss, which is the summation of these six losses. The formula for one of the six losses (image loss given text) is provided above. The remaining five terms are defined analogously.

The goal of cross-contrastive pre-training is to align the meta, image, and text embedding vectors whose output labels are the same, and push non-matching pairs apart. The cross-contrastive learning works for fine-grained classification since the loss between (text, image) and (meta, image) helps to separate images of subcategories with very similar visual features.
\subsection{Model Fine-Tuning}
\begin{figure*}
  \centering
\includegraphics[width=\linewidth]{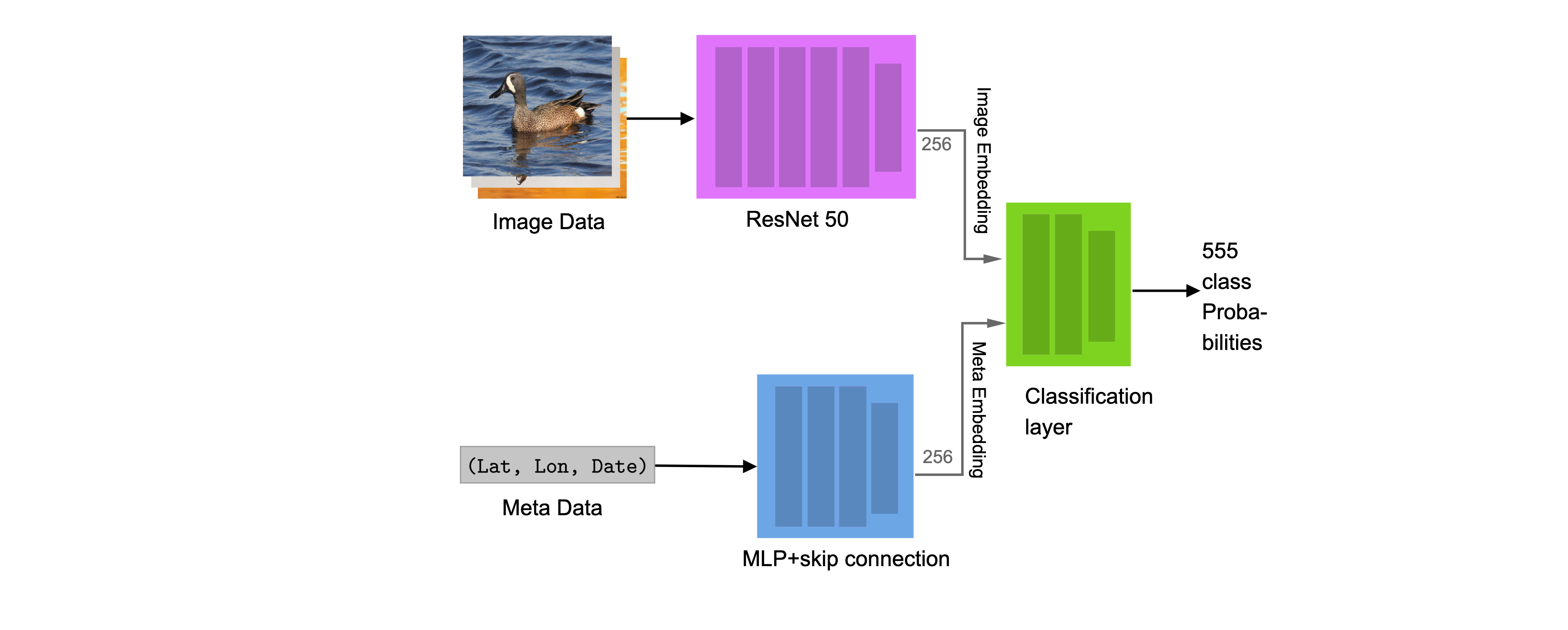}
  \caption{Proposed architecture—fine-tuning and inference after pre-training using metadata and image encoders.}
  \label{fig:infer}
\end{figure*}

\begin{figure*}
  \centering
\includegraphics[width=\linewidth]{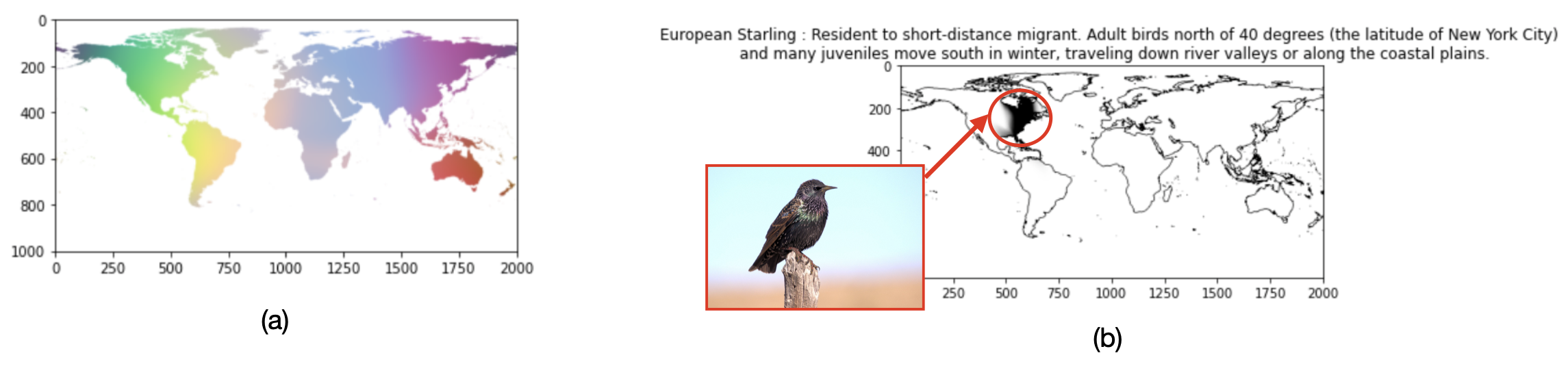}
  \caption{Qualitative Results : (a) Resulting embeddings for each input location from our model architecture trained on NABirds ~\cite{NAbird} dataset. (b) Heat map of plausible locations for the European Starling given a fixed image and mid-year date.}
  \label{fig:visualAB}
\end{figure*}

In the first step, we pre-trained the meta, image, and text encoder models using the cross-contrastive pre-training. Now, in the second step shown in Fig.~\ref{fig:infer} \label{fig:infer}, we take the output of the metadata encoder model and the ResNet-50\cite{Reset50} model and concatenate the embeddings. We use the concatenated 512-D embedding as input to a 2-layer shallow classifier that outputs 555 classes~\cite{NAbird}. We trained these models using cross-entropy loss ~\cite{Cross-Entropy-Loss} applied on the class level in contrast to the batch level while pre-training. We used Adam\cite{Adam} as an optimizer for both training procedures.
\section{Experiments}
In this section, we discuss dataset and experimentation details, followed by quantitative and qualitative results analysis.

\subsection{Dataset} 
We conduct experiments on NABirds \cite{NAbird}. It is a collection of 48,562 annotated photographs of the 555 species of birds that are commonly observed in North America. The dataset has 23,929 training images and 24,633 testing images. It contains the metadata such as latitude, longitude, and date. 

\subsection{Implementation Details}
We compared our results with a state-of-the-art model \cite{GeoPrior}. Additionally, we conducted experiments using various training settings. First, as a baseline, we trained the InceptionV3\cite{InceptionV3} model only using an image as an input. Then, we appended metadata as an extra channel on the input. We ran experiments using contrastive pre-training \cite{ContrastiveSupervised} and using the CLIP\cite{CLIP} model, considering pseudo-text and image data. 
Batch size is 32 with $\tau$ = log(0.007); training runs for 200 epochs. We pre-trained meta, text, and image encoder models through AdamW\cite{AdamW} with learning rates 5e-5, 1e-5, and 1e-4. We use Adam~\cite{Adam} for fine-tuning. Our implementation is based on PyTorch. To support future research, the code will be released after the conference.

\subsection{Quantitative Evaluation}
We report top-1 accuracy after 200 epochs. Our method reaches 84.44\%, outperforming Geo-Prior~\cite{GeoPrior} (81.50\%) and other metadata-based variants. Compared with our vision-only baseline, the gain is +7.83\%.

To isolate the effect of contrastive pre-training, we keep the backbone and hyperparameters fixed and compare against a supervised contrastive setup~\cite{ContrastiveSupervised}. Contrastive pre-training yields a clear improvement.

\begin{table}[t]
\centering
\begin{tabular}{lcc}
\toprule
\textbf{Model} & \textbf{Metadata} & \textbf{Top-1 (\%) Mean $\pm$ SD (8 runs)} \\
\midrule
InceptionV3~\cite{InceptionV3} & $\times$ & 76.61 $\pm$ 0.49 \\
InceptionV3 + Metadata & $\checkmark$ & 77.56 $\pm$ 0.46 \\
\makecell[l]{InceptionV3\\+ Contrastive Learning~\cite{ContrastiveSupervised}} & $\times$ & 79.84 $\pm$ 0.51 \\
CLIP (ResNet-50)~\cite{CLIP} & $\times$ & 81.66 $\pm$ 0.44 \\
GeoPrior Model~\cite{GeoPrior} & $\checkmark$ & 81.50 $\pm$ 0.52 \\
\textbf{Ours (6-term objective)} & $\checkmark$ & \textbf{84.44 $\pm$ 0.37} \\
\bottomrule
\end{tabular}
\vspace{4pt}
\caption{Mean $\pm$ standard deviation over eight independent runs on NABirds.}
\label{tab:results}
\end{table}
To account for stochastic training variation, we repeated each experiment across eight independent runs with different random seeds and report the mean $\pm$ standard deviation of top-1 accuracy in Table~\ref{tab:results}. The low variance ($\pm 0.37\%$ for our model) demonstrates stable convergence and strong reproducibility. Notably, relative to a CLIP-style two-term objective that uses only image–text losses, our six-term objective with metadata yields an additional $+2.78\%$ absolute improvement in top-1 accuracy.

\paragraph{Ablation of Cross-Contrastive Losses.}
Removing one or more terms from the six-part cross-contrastive objective weakens the transitive coupling among modalities and degrades both alignment and accuracy. Each loss direction (e.g., image$\leftrightarrow$metadata, image$\leftrightarrow$text) anchors complementary semantics in the shared embedding space; dropping even a single term breaks geometric consistency and induces partial modality drift. For example, omitting $L_{I,M}$ eliminates spatial constraints on appearance features, leading to confused embeddings for visually similar classes, while removing $L_{T,M}$ decouples textual priors from geo-temporal cues, flattening the embedding topology and reducing class separability. Across ablations, these omissions typically yield a 1–3\% drop in top-1 accuracy and produce visibly less coherent clusters—evidence that the full six-term objective is essential for stable tri-modal alignment and fine-grained discrimination.

\subsection{Qualitative Evaluation}
To understand the impact of metadata on overall performance, we plot Fig.~\ref{fig:visualAB}(b) the probability value of a specific class for each location (longitude and latitude) keeping the input image constant with the constant date (mid of the year). The heat map highlights plausible regions for the European Starling, indicating that the model captures object–location relationships.
Our architecture captures the relationship between objects and locations. Fig.~\ref{fig:visualAB}(a) illustrates the resulting embeddings for each input location from our model trained on NABirds \cite{NAbird} dataset. By applying the embedding function to each location, we can generate a feature vector embedding. After that, we use ICA \cite{hyvarinen2000independent} to project the embedded features to three-dimensional space and mask out the ocean for visualization.

\section{Conclusion and Future Work}

We introduced a unified framework for fine-grained visual classification that jointly models images, text, and spatio-temporal metadata via cross-contrastive pre-training. By aligning all three modalities in a shared embedding space and fine-tuning a lightweight classifier, our approach achieves 84.44\% top-1 accuracy on NABirds~\cite{NAbird}, outperforming strong baselines and underscoring the value of geo-temporal context for disambiguating visually similar species.

Our approach still has limits: it depends on having reliable metadata and adds pre-training cost compared to vision-only setups. Next, we plan to (i) scale to larger and more diverse datasets (e.g., iNaturalist), (ii) run careful ablations to measure how much each modality and loss term helps, and (iii) build stronger metadata encoders that include ecological signals such as habitat, elevation, and seasonality. The same idea may also help in areas like medical imaging and remote sensing, where context is important at inference time.


%

\bibliographystyle{IEEEtran}
\bibliography{sample}

\begin{thebibliography}{10}
\providecommand{\url}[1]{#1}
\csname url@samestyle\endcsname
\providecommand{\newblock}{\relax}
\providecommand{\bibinfo}[2]{#2}
\providecommand{\BIBentrySTDinterwordspacing}{\spaceskip=0pt\relax}
\providecommand{\BIBentryALTinterwordstretchfactor}{4}
\providecommand{\BIBentryALTinterwordspacing}{\spaceskip=\fontdimen2\font plus
\BIBentryALTinterwordstretchfactor\fontdimen3\font minus \fontdimen4\font\relax}
\providecommand{\BIBforeignlanguage}[2]{{%
\expandafter\ifx\csname l@#1\endcsname\relax
\typeout{** WARNING: IEEEtran.bst: No hyphenation pattern has been}%
\typeout{** loaded for the language `#1'. Using the pattern for}%
\typeout{** the default language instead.}%
\else
\language=\csname l@#1\endcsname
\fi
#2}}
\providecommand{\BIBdecl}{\relax}
\BIBdecl

\bibitem{NAbird}
G.~Van~Horn, S.~Branson, R.~Farrell, S.~Haber, J.~Barry, P.~Ipeirotis, P.~Perona, and S.~Belongie, ``Building a bird recognition app and large scale dataset with citizen scientists: The fine print in fine-grained dataset collection,'' in \emph{Proceedings of the IEEE Conference on Computer Vision and Pattern Recognition}, 2015, pp. 595--604.

\bibitem{SigLIP2_2025}
M.~Tschannen \emph{et~al.}, ``Siglip 2: Multilingual vision-language encoders with improved semantic understanding, localization, and dense features,'' \emph{arXiv}, 2025.

\bibitem{GeoCLIP_2023}
V.~Vivanco~Cepeda, N.~Nayak, and M.~Shah, ``Geoclip: Clip-inspired alignment between locations and images for effective worldwide geo-localization,'' in \emph{NeurIPS}, 2023.

\bibitem{RANGE_CVPR_2025}
A.~Dhakal \emph{et~al.}, ``{RANGE}: Retrieval augmented neural fields for multi-resolution geo-embeddings,'' in \emph{CVPR}, 2025.

\bibitem{Brun_NatComm_2024}
P.~Brun \emph{et~al.}, ``Multispecies deep learning using citizen science data improves fine-grained spatiotemporal biodiversity mapping,'' \emph{Nature Communications}, 2024.

\bibitem{GeoPriors_2025}
A.~Zhu, C.~Lange, and M.~Hamilton, ``Investigating different geo priors for image classification,'' \emph{arXiv}, 2025.

\bibitem{multiattention-CNN}
H.~Zheng, J.~Fu, T.~Mei, and J.~Luo, ``Learning multi-attention convolutional neural network for fine-grained image recognition,'' in \emph{Proceedings of the IEEE International Conference on Computer Vision (ICCV)}, Oct 2017.

\bibitem{Ding_2019_ICCV}
Y.~Ding, Y.~Zhou, Y.~Zhu, Q.~Ye, and J.~Jiao, ``Selective sparse sampling for fine-grained image recognition,'' in \emph{Proceedings of the IEEE/CVF International Conference on Computer Vision (ICCV)}, October 2019.

\bibitem{zheng2019looking}
H.~Zheng, J.~Fu, Z.-J. Zha, and J.~Luo, ``Looking for the devil in the details: Learning trilinear attention sampling network for fine-grained image recognition,'' in \emph{Proceedings of the IEEE/CVF Conference on Computer Vision and Pattern Recognition}, 2019, pp. 5012--5021.

\bibitem{LearningtoNavigate}
\BIBentryALTinterwordspacing
Z.~Yang, T.~Luo, D.~Wang, Z.~Hu, J.~Gao, and L.~Wang, ``Learning to navigate for fine-grained classification,'' 2018. [Online]. Available: \url{https://arxiv.org/abs/1809.00287}
\BIBentrySTDinterwordspacing

\bibitem{attentive-pairwise-interaction}
P.~Zhuang, Y.~Wang, and Y.~Qiao, ``Learning attentive pairwise interaction for fine-grained classification,'' in \emph{Proceedings of the AAAI Conference on Artificial Intelligence}, vol.~34, no.~07, 2020, pp. 13\,130--13\,137.

\bibitem{BilinearCNNs}
\BIBentryALTinterwordspacing
T.-Y. Lin, A.~RoyChowdhury, and S.~Maji, ``Bilinear cnns for fine-grained visual recognition,'' 2015. [Online]. Available: \url{https://arxiv.org/abs/1504.07889}
\BIBentrySTDinterwordspacing

\bibitem{bilinear-p0oling}
C.~Yu, X.~Zhao, Q.~Zheng, P.~Zhang, and X.~You, ``Hierarchical bilinear pooling for fine-grained visual recognition,'' in \emph{Proceedings of the European conference on computer vision (ECCV)}, 2018, pp. 574--589.

\bibitem{deep-bilinear-transformation}
H.~Zheng, J.~Fu, Z.-J. Zha, and J.~Luo, ``Learning deep bilinear transformation for fine-grained image representation,'' \emph{Advances in Neural Information Processing Systems}, vol.~32, 2019.

\bibitem{weakly-complementary-models}
W.~Ge, X.~Lin, and Y.~Yu, ``Weakly supervised complementary parts models for fine-grained image classification from the bottom up,'' in \emph{Proceedings of the IEEE/CVF Conference on Computer Vision and Pattern Recognition}, 2019, pp. 3034--3043.

\bibitem{Reset50}
K.~He, X.~Zhang, S.~Ren, and J.~Sun, ``Deep residual learning for image recognition,'' 2015.

\bibitem{ContrastiveSupervised}
P.~Khosla, P.~Teterwak, C.~Wang, A.~Sarna, Y.~Tian, P.~Isola, A.~Maschinot, C.~Liu, and D.~Krishnan, ``Supervised contrastive learning,'' \emph{Advances in Neural Information Processing Systems}, vol.~33, pp. 18\,661--18\,673, 2020.

\bibitem{CLIP}
A.~Radford, J.~W. Kim, C.~Hallacy, A.~Ramesh, G.~Goh, S.~Agarwal, G.~Sastry, A.~Askell, P.~Mishkin, J.~Clark \emph{et~al.}, ``Learning transferable visual models from natural language supervision,'' in \emph{International Conference on Machine Learning}.\hskip 1em plus 0.5em minus 0.4em\relax PMLR, 2021, pp. 8748--8763.

\bibitem{combine-vision-language}
\BIBentryALTinterwordspacing
X.~He and Y.~Peng, ``Fine-grained image classification via combining vision and language,'' in \emph{2017 {IEEE} Conference on Computer Vision and Pattern Recognition ({CVPR})}.\hskip 1em plus 0.5em minus 0.4em\relax {IEEE}, jul 2017. [Online]. Available: \url{https://doi.org/10.1109%2Fcvpr.2017.775}
\BIBentrySTDinterwordspacing

\bibitem{DistilBert}
\BIBentryALTinterwordspacing
V.~Sanh, L.~Debut, J.~Chaumond, and T.~Wolf, ``Distilbert, a distilled version of bert: smaller, faster, cheaper and lighter,'' 2019. [Online]. Available: \url{https://arxiv.org/abs/1910.01108}
\BIBentrySTDinterwordspacing

\bibitem{Cross-Entropy-Loss}
\BIBentryALTinterwordspacing
Z.~Zhang and M.~R. Sabuncu, ``Generalized cross entropy loss for training deep neural networks with noisy labels,'' 2018. [Online]. Available: \url{https://arxiv.org/abs/1805.07836}
\BIBentrySTDinterwordspacing

\bibitem{Adam}
\BIBentryALTinterwordspacing
D.~P. Kingma and J.~Ba, ``Adam: A method for stochastic optimization,'' 2014. [Online]. Available: \url{https://arxiv.org/abs/1412.6980}
\BIBentrySTDinterwordspacing

\bibitem{GeoPrior}
\BIBentryALTinterwordspacing
O.~Mac~Aodha, E.~Cole, and P.~Perona, ``Presence-only geographical priors for fine-grained image classification,'' 2019. [Online]. Available: \url{https://arxiv.org/abs/1906.05272}
\BIBentrySTDinterwordspacing

\bibitem{InceptionV3}
\BIBentryALTinterwordspacing
C.~Szegedy, V.~Vanhoucke, S.~Ioffe, J.~Shlens, and Z.~Wojna, ``Rethinking the inception architecture for computer vision,'' 2015. [Online]. Available: \url{https://arxiv.org/abs/1512.00567}
\BIBentrySTDinterwordspacing

\bibitem{AdamW}
\BIBentryALTinterwordspacing
I.~Loshchilov and F.~Hutter, ``Decoupled weight decay regularization,'' 2017. [Online]. Available: \url{https://arxiv.org/abs/1711.05101}
\BIBentrySTDinterwordspacing

\bibitem{hyvarinen2000independent}
A.~Hyv{\"a}rinen and E.~Oja, ``Independent component analysis: algorithms and applications,'' \emph{Neural networks}, vol.~13, no. 4-5, pp. 411--430, 2000.

\end{thebibliography}

\end{document}